\documentclass[10pt,twocolumn,letterpaper]{article}

\usepackage{iccv}
\usepackage{times}
\usepackage{epsfig}
\usepackage{graphicx}
\usepackage{amsmath}
\usepackage{amssymb}
\usepackage{booktabs}
\usepackage{multirow}
\usepackage[table,xcdraw]{xcolor}
\usepackage{subcaption}

\usepackage[pagebackref=true,breaklinks=true,letterpaper=true,colorlinks,bookmarks=false]{hyperref}

\iccvfinalcopy 

\def\iccvPaperID{}

\ificcvfinal\fi

\begin{document}

\title{FaceFusion: Exploiting Full Spectrum of Multiple Datasets}

\author{Chiyoung Song, Dongjae Lee\\
Naver Cloud\\
{\tt\small \{chiyoung.song, dongjae.lee\}@navercorp.com}
}

\maketitle
\ificcvfinal\thispagestyle{empty}\fi

\begin{abstract}
The size of training dataset is known to be among the most dominating aspects of training high-performance face recognition embedding model. Building a large dataset from scratch could be cumbersome and time-intensive, while combining multiple already-built datasets poses the risk of introducing large amount of label noise. We present a novel training method, named FaceFusion. It creates a fused view of different datasets that is untainted by identity conflicts, while concurrently training an embedding network using the view in an end-to-end fashion. Using the unified view of combined datasets enables the embedding network to be trained against the entire spectrum of the datasets, leading to a noticeable performance boost. Extensive experiments confirm superiority of our method, whose performance in public evaluation datasets surpasses not only that of using a single training dataset, but also that of previously known methods under various training circumstances.
\end{abstract}
\section{Introduction}

The priority focus of modern face recognition studies has been in-line with that of representation learning studies: amplifying the representative power of embedding vectors within the feature space. With continued development and refinement of various training methods, face recognition models have seen significant improvement in terms of evaluation accuracy in recent years. However, equally important is the recent advent of publicly available large-scale facial image datasets. These large datasets have generally been curated by either partially or entirely automated crawling of publicly available facial images, followed by different approaches of clustering the images by their identities. 

While the approach of building such large datasets from scratch have been studied widely, considerably less amount of attention has been given to the methods of starting from already-curated datasets. Combining different datasets is naturally beneficial from the fact that they have already gone through some degree of refinements that include identity-wise clustering or noise removal, but because their image sources generally come from public datasets of celebrities or random web crawling, careful consideration is required to handle conflicting identities. Identity conflict is destructive to the overall model performance because physically same identities are interpreted as distinct identities, which the model is incorrectly taught to distinguish. A trivial solution to this inter-class label noise would be to train a rectifier model \cite{9710478} to adjust such identities, but such solution would be heavily dependent on training a robust rectifier model. Another possible approach is explored in \cite{liu2021switchable} as a posterior data cleaning process, but it potentially requires \(O(MN)\) memory space, where \(M\) is the number of conflicting identities, and \(N\) is the number of datasets. DAIL~\cite{9412121}, to our best knowledge, is the only work that studies the way of using multiple datasets concurrently without a separate dataset-manipulation process by introducing dataset-aware softmax. While the approach of DAIL somewhat mitigates the conflicting identity problem, we argue that the performance gain from DAIL is suboptimal, because dataset-aware softmax essentially isolates the representations learned from each dataset from the rest of the datasets. This reduces the scope of final softmax operation from the entire embedding space to the subspaces fitted to each dataset, preventing the embedding model to reach global optimization.\newline
\hspace*{1em} To suppress the inevitably introduced inter-class label noise when combining different datasets, and to further improve from the limited performance gain of DAIL, we introduce \textit{FaceFusion}. Our approach begins from the method of DAIL, and after the model parameters are stabilized, observed as the well-known \textit{slow-drift}~\cite{2020slowdrift} phenomenon, it directly \textit{fuses} different datasets into a unified global embedding space, while also merging the class proxies of conflicting identities. Doing so effectively enables viewing multiple datasets as a unified dataset, which the embedding model can exploit to expand its optimization scope from within each dataset to the whole datasets, resulting in superior performance. Extensive experiments confirms that FaceFusion outperforms not only the models trained using single dataset, but also the models trained with multiple datasets either by naive concatenation or by the method of DAIL. We further prove that FaceFusion maintains its superiority over the aforementioned methods under varying severity of conflicting identities of each dataset, ranging from completely disjoint to vastly overlapping with each other.

\section{Related Works}

\subsection{Face Recognition Datasets}

The performance of face recognition model has been observed to be roughly proportional to the size of training dataset, which has encouraged series of studies and efforts~\cite{yi2014learning,cao2018celeb,guo2016ms,wang2018devil} on creating larger and larger datasets. CASIA-Webface~\cite{yi2014learning} is created by crawling and annotating the images of celebrities registered on the IMDb~\cite{IMDb}. CelebA~\cite{cao2018celeb} stems from CelebFaces~\cite{sun2013hybrid, sun2014deep}, whose source data is also collected from the web with the names of celebrities as queries. MS-Celeb-1M~\cite{guo2016ms} is also created by collecting images of celebrities from freebase. Likewise, VGGFace2~\cite{cao2018vggface2} is built by sourcing the list of id from freebase and collecting images through Google image search. As larger datasets generally contribute positively to the model performance, exploring the ways of combining the already-built datasets to make a new larger one may also be beneficial. However, such direction has been given inadequate amount of attention.

\subsection{Face Recognition Loss}

After the early attempts of triplet-based~\cite{2015triplet} loss, the focus of face recognition studies has shifted toward margin-based softmax losses~\cite{2018amsoftmax, 2017sphereface, 2019arcface, wang2018cosface}. These methods study various ways of applying margins, including angular additive, multiplicative, and geodesic additive, to the positive pair comparisons in order to further encourage intra-class compactness of the embedding model. Dynamically adjusting the magnitude of margins is studied in~\cite{2019adacos, 2020curricular}. Making the class proxies evolve along with the embedding network is discussed in~\cite{9577336}. Approaches other than softmax-based have also been actively explored. CircleLoss~\cite{circleloss} proposes methods to smooth the decision boundaries, and the ways to apply it outside softmax-like structures. A successor to~\cite{2017sphereface}, 
SphereFace2~\cite{y2021sphereface2} seeks to prove that binary cross entropy between the embedding feature and class proxies suffices model optimization.

\subsection{Face Recognition under Label Noise}
Training noise-robust embedding network has been explored by numerous studies. 
Some of the early approaches~\cite{2017ding, 2018wu} aim to make the network less susceptible to noise by augmenting the network structures with additional operations. Assigning the noisy samples to auxiliary class proxies to isolate their negative effect has been proposed by~\cite{liu2021switchable, subcenterarc}.
PFC~\cite{2022pfc} primarily focuses on training efficiency, but by randomly sampling from the whole class proxies for negative pair generation it also reduces the impact of noisy sample.
Dynamic Training Data Dropout is introduced in~\cite{9591391} that filters out unstable samples as the training progresses.
A meta-learning approach is explored in~\cite{9710478} by partitioning the datasets into meta-train and meta-test to train a separate cleaner model.
Exploiting the long-tail noisy data distribution is studied in~\cite{8954137} by putting more emphasis on the samples from head distribution with a newly designed loss function that dynamically focuses on either the model prediction or the given label.
A semi-supervised training approach of~\cite{2022Liu} is to repeatedly assign pseudo-labels to unlabeld sample and drop unreliable samples with multi-agent data exchange.
Dynamically weighing each sample according to the position of cosine similarity against the class proxy within the overall similarity distribution is explored in~\cite{8953947}.

It is noteworthy to mention that the aforementioned works explore the ways to weaken the effect of label noise in single-dataset paradigm, which is naturally expected to have relatively controlled amount of noise. Our work focuses on combining multiple datasets, which accompanies significantly severe amount of label noise from conflicting identities.

\graphicspath{{./figures/}}
\section{Methods}
\begin{figure*}[t]
    \centering
    \includegraphics[width=\textwidth]{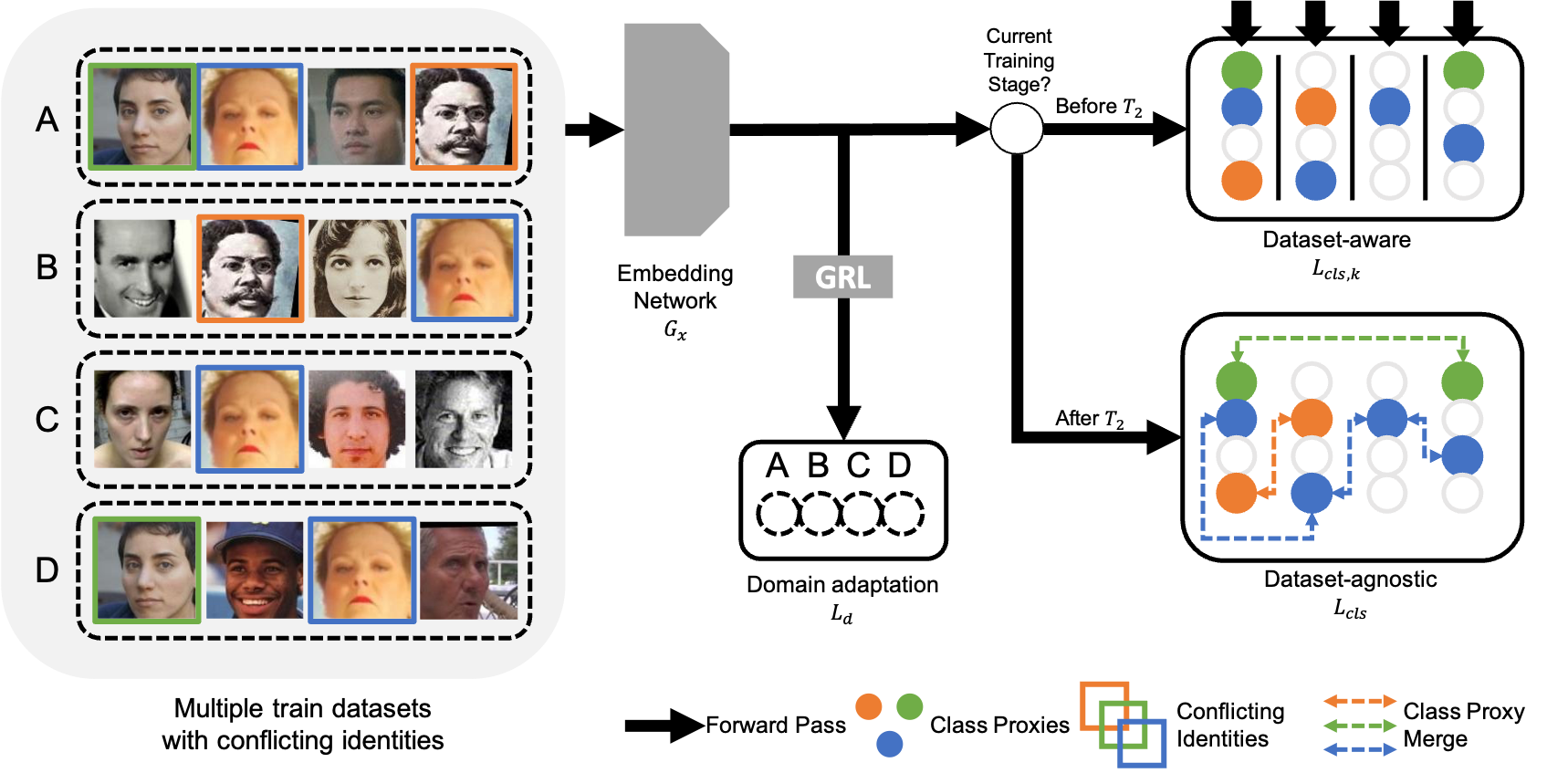}
    \caption{Overview of FaceFusion. \(L_{cls, k}\) and \(L_{cls}\) shares the same class proxies. While \(L_{cls, k}\) limits the softmax calculations to each dataset, \(L_{cls}\) merges the class proxies of same identity, and removes the barriers between the datasets. GRL reverses the direction of gradients to encourage generalization of embedding features.}
    \label{fig:overview}
\end{figure*}

Through the following subsections, we discuss DAIL~\cite{9412121}, and how it achieves suboptimal performance under multiple training dataset setting. From there, we present FaceFusion, which effectively removes the architectural weakness of DAIL and consequently results in superior performance when trained using multiple datasets.

\subsection{Preliminary}
Currently, the state-of-the-art methods~\cite{2018amsoftmax,2017sphereface,wang2018cosface,2019arcface,subcenterarc} to train high-performance face recognition models are dominantly formulated as variants of angular softmax loss
\begin{equation}\label{1}
 L_{cls} = \frac{1}{N}\sum_{i=1}^{N}{-\log{\frac{e^{\cos\theta_{y_i}  }}{e^{\cos\theta_{y_i}} + \sum_{j \neq y_i}^{C}{e^{\cos\theta_{j}}}}}},
\end{equation}
where \(\theta_{{y_i}}\) and \(\theta_{j}\) represent the angular distances between \(i^{th}\) sample and its positive class proxy \(\vec{W}_{y_{y_i}}\) and a negative class proxy \(\vec{W}_{y_{j}}\), respectively.

As discussed in~\cite{liu2021switchable}, the overall training objective of face recognition models is in line with that of representation learning: to compactly pack the features originating from the samples of the same label, and to spread the features of different labels as much as possible within the feature space. A considerable number of works have evolved to use additional angular margins~\cite{2018amsoftmax,2017sphereface,wang2018cosface,2019arcface,2019adacos,circleloss,y2021sphereface2} when computing the logits in order to further encourage the intra-class compactness and inter-class spread, and report improved performances. 

However, trivially applying Eq.\ref{1} may lead to severe performance degradation if the uniqueness of each identity is not guaranteed. This inter-class noise, where the same individual is categorized multiple times under different labels, is of a disturbance to the softmax training of encouraging intra-class compactness and inter-class spread~\cite{subcenterarc,liu2021switchable, 9412121}, because the flow of gradients from optimizing \(\theta_{j}\) and \(\theta_{{y_i}}\) may conflict. This inter-class noise problem becomes more evident if more than one dataset is being used without a proper label adjustments. As many modern public datasets for face recognition take their sources from either celebrity database or web crawlings, assuring that any two public datasets having zero conflicting identities becomes implausible. Likewise, it becomes nearly intractable to correctly rectify each and every one of the conflicting identities from any two datasets without specific meta-labels.

\subsection{Dataset Aware and Invariant Learning}
DAIL~\cite{9412121} avoids this inter-class noise problem under multiple dataset settings by limiting the softmax formulation to each dataset as
\begin{equation}\label{dail_selector}
{1_{k_{j}={k_{y_{i}}}}} = \begin{cases} 
      1 & {k_{j}={k_{y_{i}}}} \\
      0 & otherwise 
   \end{cases}
\end{equation}
\begin{equation}\label{softmaax_dail}
 L_{cls,k} = \frac{1}{N}\sum_{i=1}^{N}{-\log{\frac{e^{\cos\theta_{y_i}  }}{e^{\cos\theta_{y_i}} + \sum_{j \neq y_i}^{C}{{1_{k_{j}={k_{y_{i}}}}}e^{\cos\theta_{j}}}}}},
\end{equation}
where \(k_{j}\) and \(k_{y_i}\) are the source datasets of class \(j\) and \(y_i\) respectively. By this isolation of softmax calculation, DAIL prevents the issue of incorrectly using an identity as both positive and negative pairs if the identity appears in more than one datasets. We argue that, while doing so suffices to reduce the effect of inter-class noise, it fails to take the full benefits of using multiple datasets to explore a broader spectrum of identities. With Eq.\ref{softmaax_dail}, an output feature from a sample is compared only against the class proxies of the same dataset, effectively losing the opportunity to be further optimized globally by considering all the datasets as a whole. A preliminary mitigation to this is briefly studied in \cite{9412121} with random sampling, but such attempt is insufficient in suppressing the label noise entirely. We now propose FaceFusion, which can effectively alleviate this limitation of~\cite{9412121} by fusing the datasets into one while unobtrusively preserving the samples of overlapping identities.

\subsection{FaceFusion}

Our approach stems from~\cite{9412121}, where each dataset is used concurrently yet kept in isolation from one another. We observe that {\textit{slow-drift}}~\cite{2020slowdrift} also applies  under face recognition paradigm, which in consequence stabilizes the class proxies. This leads to the formulation of class proxies of each dataset in relatively stable yet intertwined hyperplanes sharing common anchor points via the class proxies of conflicting identities. Additionally optimized with a separate domain adaptation loss~\cite{9412121}, possible discrepancies within conflicting class proxies arising from the domain gap are minimized. Therefore, we let the training using dataset-aware softmax~\cite{9412121} continue for the first \(T_2\) proportions of the total training steps to stabilize the hyperplanes.

Once the optimizations of class proxies within each dataset have matured, we adopt the posterior data cleaning strategy~\cite{liu2021switchable} to merge the class proxies \(\vec{W}_{y_{i}}\) and \(\vec{W}_{y_{j}}\) if their similarity, calculated as 
\begin{equation}\label{4}
sim(\vec{W}_{y_{i}}, \vec{W}_{y_{j}}) = \frac{\langle {\vec{W}_{y_{i}}, {\vec{W}_{y_{j}}} \rangle}}{\|\vec{W}_{y_{i}}\|_2\|\vec{W}_{y_{j}}\|_2},
\end{equation}
exceeds \(T_1\). Unlike using a separate rectifier model for such task, it becomes nearly effortless under FaceFusion, because all the class proxies are already known to the embedding network, thus making Eq.\ref{4} more reliable. This merging process essentially enables the samples of conflicting identities to be trained with a unified class proxy, and lifts the need of isolating the softmax calculation. We exploit to switch from the dataset-aware softmax~\cite{9412121} to dataset-agnostic softmax and further optimize the embedding network against the whole span of the datasets. It is important to note, however, that the domain adaptation loss~\cite{9412121} is kept for the entire duration of the training process to further regularize the embedding model.

The overview of our proposed method is presented in Figure \ref{fig:overview}. Our method is compatible to any of the single-proxy softmax-based losses~\cite{wang2018cosface,2019arcface,2017sphereface,2018amsoftmax}, or multiple-proxy based losses~\cite{liu2021switchable, subcenterarc}. For our implementation, we employ Arcface~\cite{2019arcface} loss.
\section{Experiments}

\subsection{Datasets}\label{datasets}
In this paper, we use 6 datasets for training, including Asian-Celeb~\cite{DeepGlint}, CASIA-WebFace~\cite{yi2014learning}, CelebA~\cite{cao2018celeb}, DeepGlint~\cite{DeepGlint},  MS1M~\cite{deng2019arcface}, a semi-automatically cleaned version of MS-Celeb-1M~\cite{guo2016ms} and VGGFace2~\cite{cao2018vggface2}, all of whose details are shown in Table \ref{table:dataset}. We crop, align, and resize face images to make it 112x112 pixel size as done in ~\cite{2019arcface, liu2021switchable}. For quantitative analysis of model performance, we report the accuracy against widely used evaluation sets, including LFW~\cite{huang2008labeled}, CFP-FP, CFP-FF~\cite{sengupta2016frontal}, AgedB30\cite{moschoglou2017agedb}, CALFW~\cite{zheng2017cross} and CPLFW~\cite{zheng2018cross}.

\begin{table}[h]
\begin{center}
\begin{tabular}{@{}c|cc@{}}
\toprule
\textbf{Dataset} & \textbf{\#Identity} & \textbf{\#Image} \\ \midrule
Asian-Celeb      & 94.0K                 & 2.8M             \\
CASIA-Webface    & 10.5K               & 0.5M             \\
CelebA           & 10.2K               & 0.2M             \\
DeepGlint        & 180.9K              & 6.8M             \\
MS1M             & 85.7K               & 5.8M             \\
VGGFace2         & 8.6K                & 3.1M             \\ \bottomrule
\end{tabular}
\end{center}
\caption{Statistics of data used for training.}
\label{table:dataset}
\end{table}

\subsection{Implementation Details}\label{impl_details}
All experiments are conducted with ResNet50~\cite{2019arcface}, with 512-dimensional embedding outputs. We employ most of the compatible settings of~\cite{9412121}. The total batch size is set to 1024. The weight given to the domain adaptation loss is set to 0.1. The gradient reversal layer is activated at step 80k for sections \ref{allin} and \ref{hparam}, and turned off for sections \ref{chunks} and \ref{duplicated_images}. The initial learning rate is set to 0.005, and is reduced by the factor of 10 at steps 80k, 140k, 200k, for total of 480k steps for sections \ref{allin} and \ref{hparam}. For section \ref{chunks} and \ref{duplicated_images}, the max steps and learning rate scheduling steps are halved to adjust for smaller dataset size. The Arcface~\cite{2019arcface} implementation of~\cite{insightface_git} with \(m = 0.5\) and \(s = 64\) is used throughout all experiments. SGD optimizer with momentum of 0.9 and weight decay of 5e-4 is used. The models are trained using one NVIDIA A100 GPU. PyTorch is used for experiments implementation.

\subsection{Comparison against state-of-the-arts}\label{allin}
\begin{table*}[h]
\begin{center}
\resizebox{\textwidth}{!}{\begin{tabular}{@{}cccccccccc@{}}
\toprule
\multirow{2}{*}{\textbf{Backbone}} & \multirow{2}{*}{\textbf{Trained Method}} & \multirow{2}{*}{\textbf{Trained Dataset}} & \multirow{2}{*}{\textbf{\# of ids}} & \multicolumn{6}{c}{\textbf{Evaluation Set}} \\
 &  &  &  & \textbf{LFW} & \textbf{CALFW} & \textbf{CPLFW} & \textbf{CFP-FP} & \textbf{CFP-FF} & \textbf{AgeDB30} \\ \midrule
\multirow{9}{*}{ResNet50} & \multirow{6}{*}{ArcFace} & Asian-Celeb & \multicolumn{1}{c|}{94.0K} & 99.18 & 93.32 & 82.30 & 88.83 & 99.17 & 92.57 \\
 &  & Casia-Webface & \multicolumn{1}{c|}{10.5K} & 98.18 & 89.07 & 77.82 & 87.81 & 97.29 & 86.27 \\
 &  & CelebA & \multicolumn{1}{c|}{10.2K} & 95.53 & 85.37 & 71.05 & 74.44 & 95.03 & 76.25 \\
 &  & DeepGlint & \multicolumn{1}{c|}{180.9K} & \textbf{99.70} & 95.85 & 87.95 & 92.94 & 99.69 & 97.17 \\
 &  & MS1MV2 & \multicolumn{1}{c|}{85.7K} & \textbf{99.70} & 95.67 & \textbf{89.65} & 93.97 & \textbf{99.76} & \textbf{97.28} \\
 &  & VGGFace2 & \multicolumn{1}{c|}{8.6K} & 99.47 & 93.22 & 89.48 & \textbf{94.00} & 99.27 & 92.33 \\ \cmidrule(l){2-10} 
 & ArcFace + Naive Concat & \multirow{3}{*}{Combined} & \multicolumn{1}{c|}{\multirow{3}{*}{389.9K}} & 99.67 & 95.32 & 88.6 & 94.30 & 99.60 & 96.57 \\
 & ArcFace + DAIL &  & \multicolumn{1}{c|}{} & 99.67 & 95.65 & 90.5 & 94.80 & \textbf{99.77} & 97.23 \\
 & ArcFace + FaceFusion &  & \multicolumn{1}{c|}{} & \textbf{99.70} & \textbf{95.90} & \textbf{90.92} & \textbf{95.09} & 99.74 & \textbf{97.52} \\ \bottomrule
\end{tabular}}
\end{center}
\caption{Performance comparisons against mdoels trained with single dataset, and multiple datasets. Results with Arcface method are reproduced by adopting some implementations found in~\cite{insightface_git}. Results with DAIL~\cite{9412121} are reproduced with our own implementation due to its usage of private dataset. For FaceFusion, \(T_1 = 0.7\) and  \(T_2 = 0.21\) are used. Naive concatenation method is equivalent to FaceFusion with \(T_1 > 1.0\) so that the identities are never considered for fusing at all. }
\label{table:allin}
\end{table*}

The performance comparisons of FaceFusion against different dataset combinations are given in Table \ref{table:allin}. As already has been elaborated in~\cite{9412121}, using multiple datasets is considerably advantageous over using single dataset, but there are noticeable differences among the performances of models trained with combined datasets. The model trained by naively concatenating the datasets results in even lower accuracy in some of the evaluation sets than the ones trained by using single dataset. This is well-expected, because combining the datasets without adjustments for conflicting identities inevitably generates inter-class label noises and hinders model optimization using conventional softmax-like loss functions, which is further examined in section \ref{chunks}. It is worth mentioning, however, that despite the presence of severe label-noise, naively combining multiple datasets still outperforms single-dataset models on some of the evaluation sets. We suspect that the shear number of identities obtained by combining multiple datasets overwhelms the negative effect of identity conflicts, further justifying the merit of using multiple datasets for training. Applying DAIL~\cite{9412121} improves upon the naive concatenation. FaceFusion shows superior results in most evaluation cases, proving that its use of richer softmax pool by fusing different datasets into a unified representation contributes to the global optimization of embedding model. 

\begin{figure}[h]
\begin{center}
\includegraphics[width=0.9\linewidth]{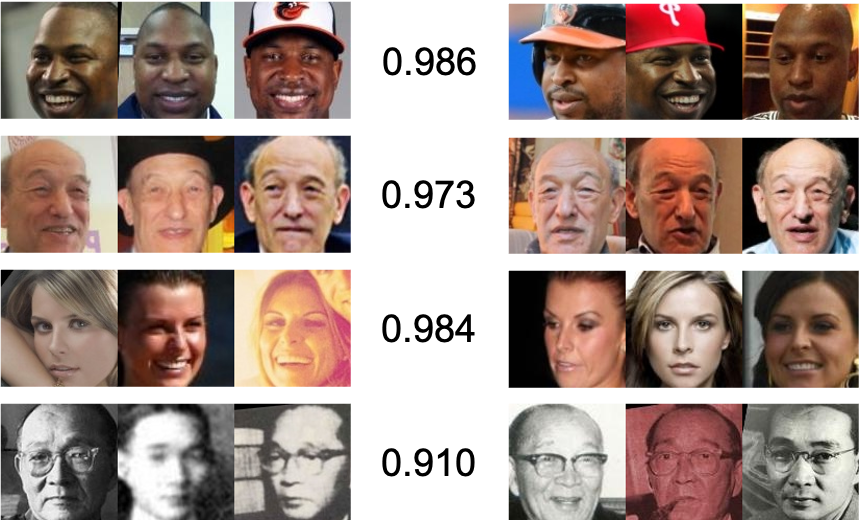}
\end{center}
\caption{Sample images of some of the merged identities, and the cosine similarity between their class proxies.}
\label{fig:merged_similarities}
\end{figure}

The datasets used in Table \ref{table:allin}, when taken into consideration as whole, consist of number of conflicting identities, as evidenced by the poor performance of the model trained by naive concatenation. FaceFusion successfully bypasses this label noise problem by the dataset-aware softmax~\cite{9412121}, and improves upon it by switching to dataset-agnostic softmax with class proxy fusion. It is critical, however, to correctly fuse the class proxies of the same identities, for incorrectly fusing different identities destructively introduces intra-class label noise. While there is no definitive ways to verify the merge correctness without referring to the extra meta-labels of each dataset, which are not generally accessible, we show examples of some of the merged identities in Figure \ref{fig:merged_similarities} that visually suggest the merged identities can safely be used for representing conflicting identities of different datasets.

\subsection{Effect of Hyperparameters}\label{hparam}
\begin{table}[h] 
\begin{center}
\begin{tabular}{@{}cc|cccc@{}}
\toprule
\multirow{2}{*}{T1} & \multirow{2}{*}{T2} & \multicolumn{4}{c}{Evaluation Set} \\
 &  & LFW & CFP-FP & \multicolumn{1}{c|}{AgeDB30} & avg. \\ \midrule
0.9 & \multirow{5}{*}{0.21} & \textbf{99.72} & 95.17 & \multicolumn{1}{c|}{97.35} & 97.41 \\
0.8 &  & 99.68 & \textbf{95.36} & \multicolumn{1}{c|}{97.35} & \textbf{97.46} \\
0.7 &  & 99.70 & 95.09 & \multicolumn{1}{c|}{\textbf{97.52}} & 97.44 \\
0.6 &  & 99.70 & 95.04 & \multicolumn{1}{c|}{97.13} & 97.29 \\
0.5 &  & 99.70 & 95.16 & \multicolumn{1}{c|}{97.27} & 97.38  \\ \midrule
\multirow{5}{*}{0.7} & 0.05 & 99.60 & 94.57 & \multicolumn{1}{c|}{96.93} & 97.03 \\
 & 0.10 & 99.67 & 94.90 & \multicolumn{1}{c|}{97.43} & 97.33 \\
 & 0.16 & \textbf{99.70} & 94.97 & \multicolumn{1}{c|}{97.27} & 97.31 \\
 & 0.21 & \textbf{99.70} & \textbf{95.09} & \multicolumn{1}{c|}{\textbf{97.52}} & \textbf{97.44} \\
 & 0.26 & \textbf{99.70} & 94.87 & \multicolumn{1}{c|}{97.15} & 97.24 \\ \bottomrule
\end{tabular}
\end{center}
\caption{Performance of FaceFusion under different \(T_1\) and \(T_2\) values. }\label{table:hparams}
\end{table}

We explore how FaceFusion behaves under different settings of \(T_1\) and \(T_2\), each governing the similarity threshold between two class proxies for merging, and the duration of parameter stabilization period, respectively. For this, we employ the same training datasets as in section \ref{allin}, but for one set of experiments we vary the values of \(T_1\) and keep \(T_2\) constant, and switch it for the other set in order to isolate the effect of each parameter. Their results are given in Table \ref{table:hparams}.

\begin{figure}[h]
\includegraphics[width=\linewidth]{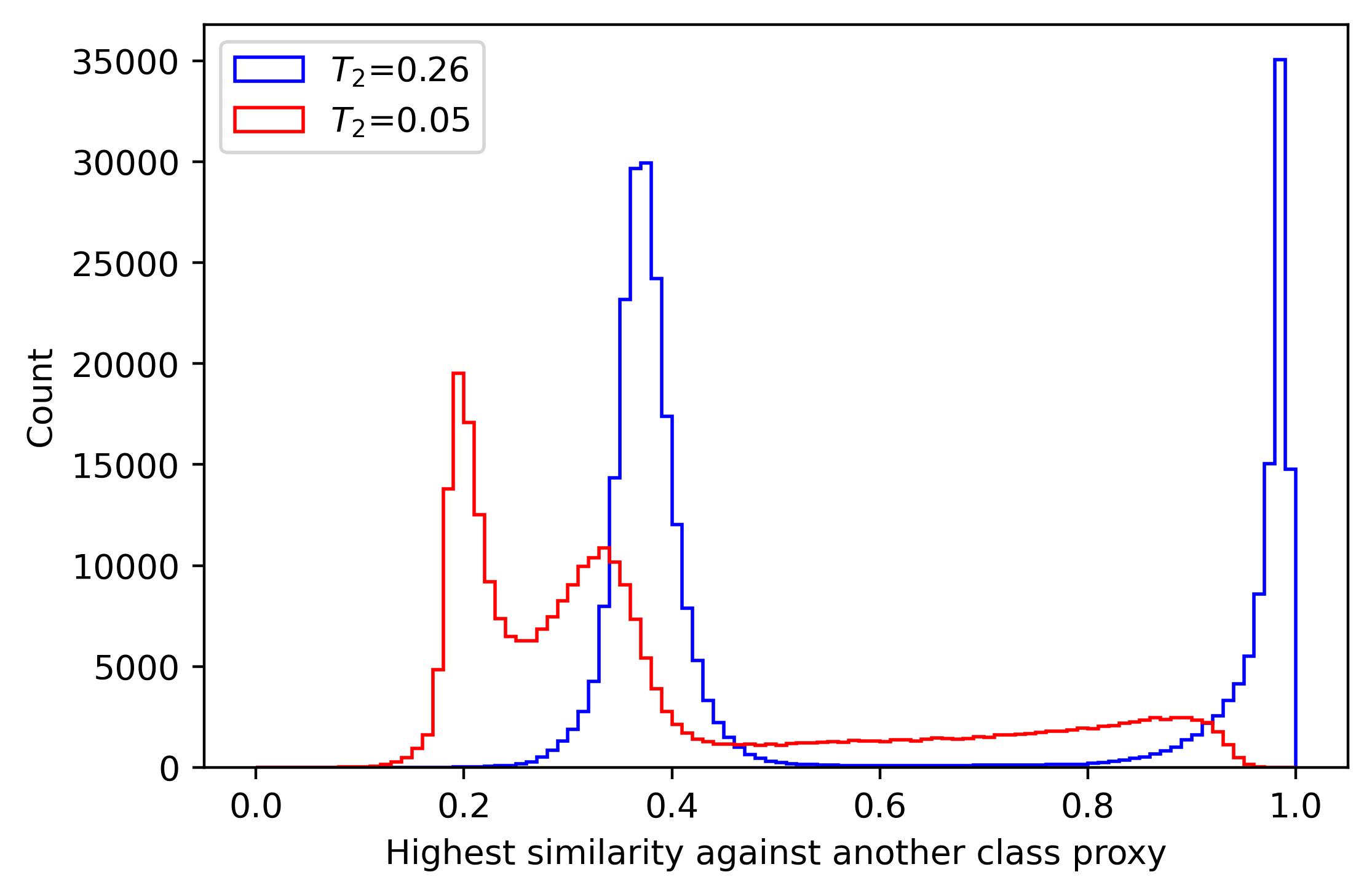}
\caption{Distribution of top-1 similarity scores for each class proxy against rest of the datasets at two different stages of the training process.}
\label{fig:sim_hist_overlay}
\end{figure}

Against our initial predictions, the effect of \(T_1\) is minimal, and hardly no relationships between the final results and the value of \(T_1\) is observed. We hypothesize that with \(T_2 = 0.21\), the class proxies of conflicting identities are so stabilized that they reside in extreme proximity to each other, causing their similarity values to climb well over our set of \(T_1\) values. To further investigate this phenomenon, we plot the distributions of similarity scores between different class proxies, as shown in Figure \ref{fig:sim_hist_overlay}. The similarity scores are observed to be grouped into two clusters, one being around 0.35, and the other being very close to 1.0. This observed distribution suggests that the effect of \(T_1\) values ranging from 0.5 to 0.8 is likely to be indistinguishable, for the highly-similar class proxy pairs are already concentrated at above 0.8, rendering \(T_1\) variation obsolete. Furthermore, this also provides an explanation to a slight performance degradation when \(T_1\) is raised from 0.8 to 0.9. Because a small portion of conflicting identities have similarity scores less than 0.9, using \(T_1 = 0.9\) and regarding them as separate identities introduces small inter-class label noise compared to \(T_1 = 0.8\). The performance degradation due to high \(T_1\) is upper-bounded by naive-concatenation result of Table \ref{table:allin}, equivalent to setting \(T_1 = 1.0\) and regarding all classes to be distinct. This suggests that FaceFusion behaves somewhat less sensitively to the values of \(T_1\), as long as sufficiently large \(T_2\) value is used to exploit from slow drift phenomenon. We acknowledge, nevertheless, that setting very low \(T_1\) values could be detrimental to the overall performance by introducing significantly large amount of intra-class label noise. 

Since using \(T_1\) as a controlled variable results in emphasizing the importance of \(T_2\), we proceed to study the impact of different  \(T_2\) values by setting \(T_1\) constant. As evidenced from the results shown in lower half rows of Table \ref{table:hparams}, FaceFusion is more susceptible to the performance loss due to inappropriate settings of \(T_2\). Although weakly monotonic with small variances, the results are shown to be proportional to the values of \(T_2\). The cause of this performance dependency on \(T_2\) is more evident when the class proxy similarities of low \(T_2\) are examined, as shown in Figure \ref{fig:sim_hist_overlay}. At this stage, the model parameters are expected to be undergoing rapid updates from their initial values, so the model has yet to learn stable class proxies for each identity, shown as the wide spread of the similarity scores. These examinations prove that waiting for the slow-drift phenomenon works favorably for identity merging, justifying our architectural decision of introducing the parameter \(T_2\) for governing the duration of model stabilizing process. The best performance is achieved with \(T_2=0.21\). The performance degradation after \(T_2=0.21\) is suspected to have resulted from the insufficient amount of optimization period given \textit{after} the classes have been merged, along with other training-related parameters as stated in section \ref{impl_details}, including the learning rate and scheduling. Although this experimentation configurations could further be tuned for better performance, we argue that such settings are be tightly dependant on the types of training datasets being used, and that the settings used are general enough for fair comparison with DAIL as well as for examining the hyperparameter effects.

\subsection{Effect of Overlapping Identities}\label{chunks} 
\begin{figure}[h]
\includegraphics[width=\linewidth]{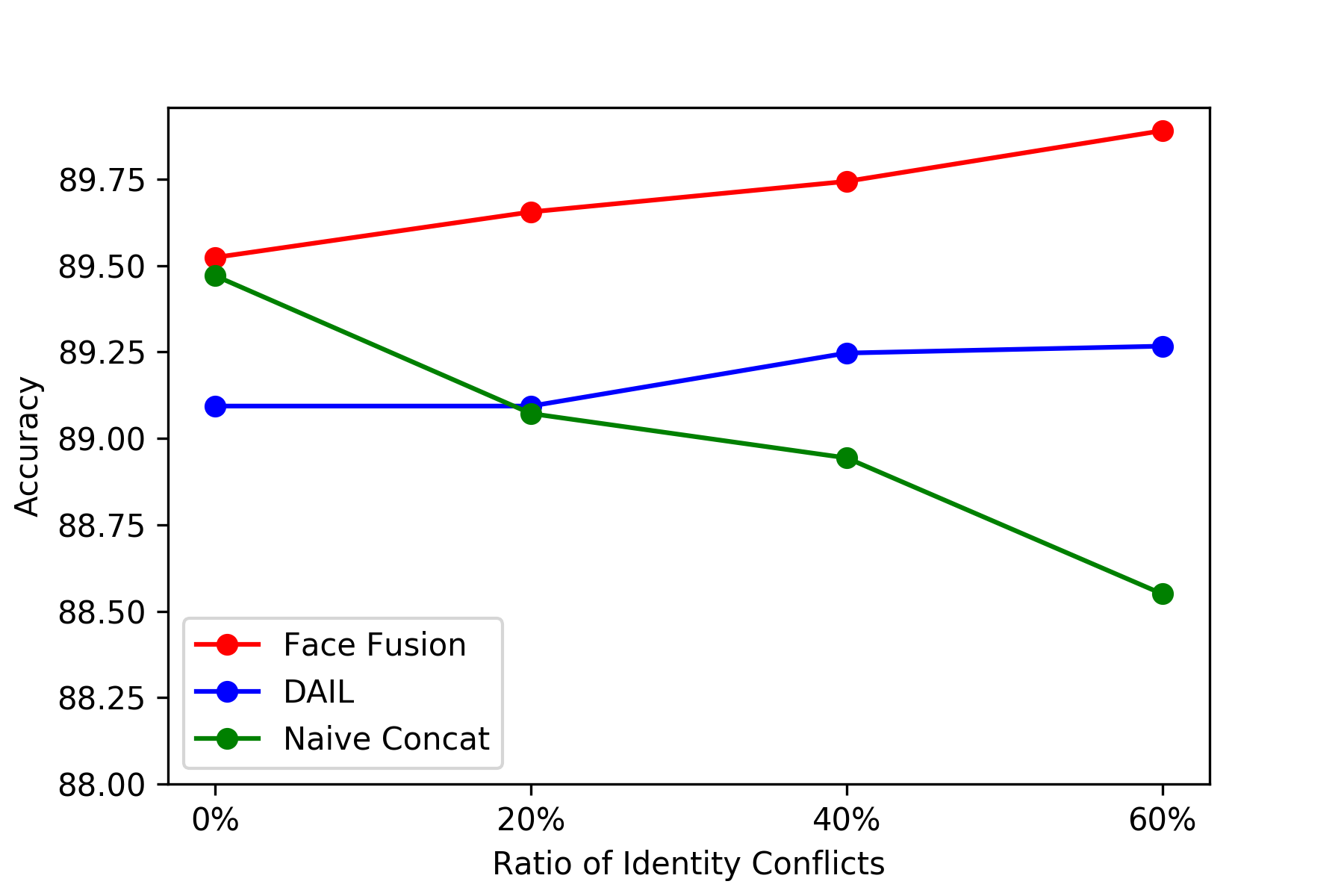}
\caption{Average accuracy of evaluation sets mentioned in ~\ref{datasets} under different \(r\). We use CASIA-Webface with \(k=8\) in this experiment. Detailed results for each evaluation set can be found in the supplementary.} 
\label{fig:chunks_acc}
\end{figure}

We conduct further experiments to study the relationship between the performance FaceFusion and various ratios of inter-class noise introduced by combining the datasets. Let \(S\) be the set of all identities in a dataset of size \(N\) identities. We evenly divide \(S\) into \(k\) different subsets, such that
\begin{equation}
\begin{aligned}
    S      &= \bigcup_{i=1}^{k}s_i, \quad \bigcap_{i=1}^{k}I_{s_i} = \emptyset \\
    N * r  &= |\bigcap_{i=1}^{k}s_i| = \sum_{i=1}^{k-1} |s_i \cap s_{i+1}| + |s_k \cap s_1|
\end{aligned}
\label{equ_subsets}
\end{equation}
are satisfied, where \(I_{s_i}\) denotes images in subset \(s_i\). We apply the configuration of Eq.\ref{equ_subsets}, along with \(k=8\), to CASIA-Webface to generate a controlled dataset \(S_{casia}\) for the experiment. The result reveals that the performance of the naive concatenation method is inversely proportional to the identity overlapping ratio, while DAIL and FaceFusion maintain their performance as shown in Figure \ref{fig:chunks_acc}. This is believed to be due to the fact FaceFusion and DAIL shares the same key advantage: the robustness against the noise introduced by conflicting identities. However, FaceFusion still outperforms DAIL by a large margin regardless of the noise ratio. This advocates the definitive advantage of FaceFusion, that it uses much larger pool of identities than DAIL for softmax calculation, whose benefit is still distinguishable even with a small-sized dataset such as CASIA-Webface. 

To closely examine this performance gap, we compare the number of negative pairs that each method uses for final softmax calculation, as shown in Table \ref{table:fc_size}. Although DAIL trains for all identities of the training dataset chunks, the final softmax calculation stays limited to each \(s_i\). This causes the performance of DAIL to fall below that of naively concatenating \(S_{casia}\) when \(r=0\), due to its far-smaller number of negative pairs. Moreover, by observing the performance of DAIL stays proportional to the size of \(s_i\) we further justify our initial assertion that increasing the number of negative pairs is beneficial for the model performance. This also validates our approach of viewing multiple datasets as a unified one for maximizing the negative pairs count. The non-equal identity counts for FaceFusion and naive concatenation when \(r=0\) are originating from the prior noise in CASIA-Webface dataset itself. FaceFusion appropriately removes the effect of this noise by merging the proxies, which may have resulted in a slight performance boost over naive concatenation. 

\begin{table}[h]
\begin{center}
\begin{tabular}{@{}c|c|c|c@{}}
\toprule
\(r\) & FaceFusion & DAIL & Naive Concat \\ \midrule
0\%  & 10560      & 1322 & 10572        \\
20\% & 10808      & 1586 & 12684        \\
40\% & 11064      & 1850 & 14796        \\
60\% & 11312      & 2114 & 16908        \\ \bottomrule
\end{tabular}
\end{center}
\caption{Number of class proxies used for the softmax calculation after \(T_2\) amount of training steps have been taken for different \(r\) of \(S_{casia}\). For DAIL, the count is equivalent to the size of each \(s_i\) in \ref{equ_subsets}. For naive concatenation, the count is equivalent to \(|S| + (N * r)\).}
\label{table:fc_size}
\label{tabel:fc_size}
\end{table}

\subsection{Effect of Image Duplications}\label{duplicated_images}
\begin{figure}[h]
\begin{center}
\includegraphics[width=\linewidth]{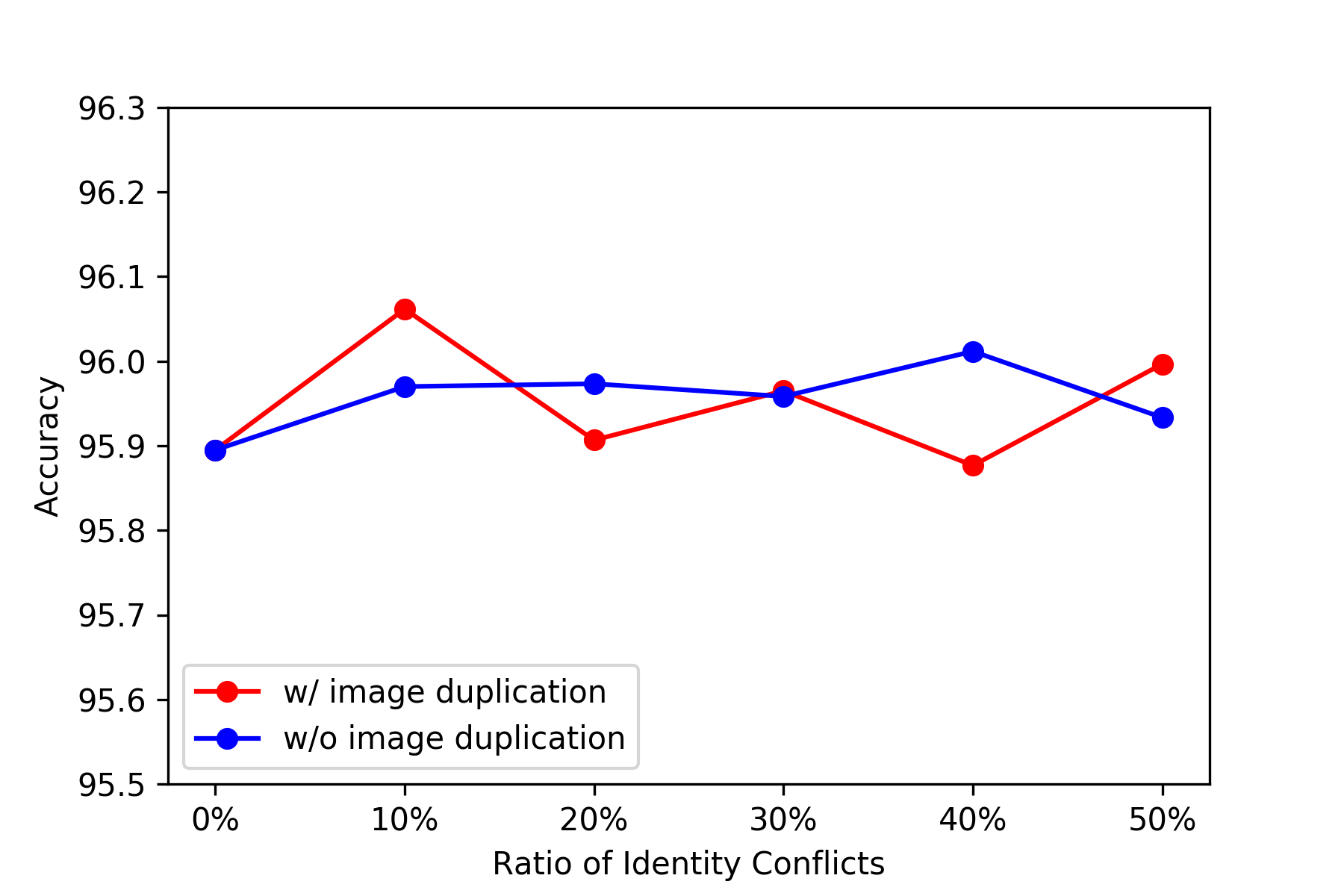}
\end{center}
\caption{Average accuracy of evaluation sets mentioned in ~\ref{datasets} under different \(r\). We use MS1M with \(k=2\) in this experiment. Detailed results for each evaluation set can be found in the supplementary.}
\label{fig:ms1m_chunks_acc}
\end{figure}

\begin{table}[h]
\begin{center}
\begin{tabular}{@{}c|c|cc@{}}
\toprule
\multirow{2}{*}{\(r\)} & \multirow{2}{*}{Conflicts} &\multicolumn{2}{c}{Merged} \\ & & Duplicated Images & Distinct Images \\ \midrule
0\% & 0 & 455 & 455 \\
10\% & 8574 & 9004 & 8725 \\
20\% & 17148 & 17514 & 17049 \\
30\% & 25722 & 26045 & 25379 \\
40\% & 34296 & 34575 & 33653 \\
50\% & 42870 & 43136 & 42036 \\ \bottomrule
\end{tabular}
\end{center}
\caption{Count of conflicting identities in \(S_{ms1m}\), and merged identities under different noise configurations \(r\) for \(S_{ms1m}\) and \(S'_{ms1m}\).}\label{table:chunk_nums}
\end{table}

Although the performance of FaceFusion is verified under varying conditions of identity conflicts, we conduct another set of experiments to observe the behavior of FaceFusion under the presence of duplicated images. For this, we choose MS1M dataset, and apply the configuration of Eq. \ref{equ_subsets} to form \(S_{ms1m}\) with \(k=2\). However, we also make a copy of \(S'_{ms1m}\) with now the subsets overlap both in terms of identities and images, so that \(\bigcap I_{s'_i} \neq \emptyset\).

We confirm, as shown in Figure~\ref{fig:ms1m_chunks_acc}, that FaceFusion behaves indifferently to image duplication regardless of the proportion of identity conflicts. Because FaceFusion conducts the identity merging of class proxies whose source features share the same embedding network, merging the class proxies trained with disjoint set of images belonging to the same identity can be reliably carried. Moreover, examining the counts of merged identities for both \(S_{ms1m}\) and \(S'_{ms1m}\), as shown in Table \ref{table:chunk_nums}, further indicates that the actual results of identity merging does not get swayed decisively by the presence of duplicating images, and no definitive connections to the actual model performance are observed. This observation further emphasizes the generalization ability of FaceFusion to more realistic environments, where the conflicts can take place in identity as well as image level. Lastly, as is the case for Table \ref{table:fc_size}, the non-zero merge counts observed for \(r=0\) are results of prior label noise in original MS1M dataset.

\section{Conclusion}
We present a novel method, \textit{FaceFusion}, of training face recognition embedding model by using multiple training datasets concurrently. FaceFusion achieves superior evaluation results over training with single dataset, as well as the previous work that attempts to solve the same issue. FaceFusion not only suppresses the negative effect of training with conflicting identities from different datasets, but also strengthens the representative power of the embedding feature by targeting the whole datasets for global optimization in a novel way, in an end-to-end fashion with no additional computation costs. Thorough experiments have proved the robustness of FaceFusion under various combinations of datasets and the number of conflicting identities.

{\small
\bibliographystyle{unsrt} 
\bibliographystyle{ieee_fullname}
\bibliography{egbib}

}

\onecolumn
\renewcommand{\thefigure}{S\arabic{figure}}
\renewcommand{\thetable}{S\arabic{table}}
\renewcommand{\thesubsection}{S\arabic{subsection}}

\section*{Supplementary Material}
\setcounter{section}{6}
\subsection{Distribution of top-1 scores}

\begin{figure}[h]
    \centering
    \subfloat[w/ image conflicts]{%
    \includegraphics[width=7cm]{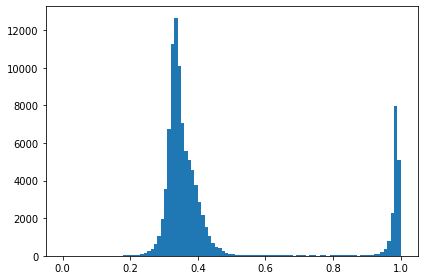}
    \label{fig:sim_img_dup}%
    }%
    \subfloat[w/o image conflicts]{%
    \includegraphics[width=7cm]{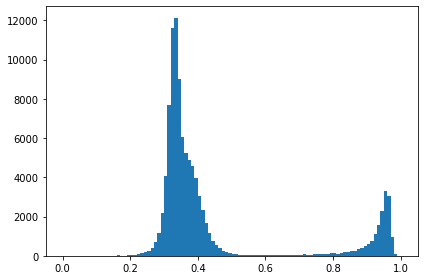}
    \label{fig:sim_img_no_dup}%
}%
\caption{Distribution of top-1 similarity scores for each class proxy against rest of the datasets under the presence of duplicated images when \(r=0.2\).} 
\label{fig:sim_distribution_appendix}
\end{figure}

Figure \ref{fig:sim_distribution_appendix} shows the distributions of top-1 similarity scores of each class proxy when \(r=0.2\) and with or without the occurrence of duplicated images. As the training process continues, the class proxies move from their randomly initialized positions to the centers of each class that are represented by the same set of images, embedded by an identical model. This makes the class proxies move extremely close to each other for the case of overlapping images, which does help merging the class proxies. Nevertheless, as discussed in Section 4.6, FaceFusion is capable of managing either cases equally well.

\subsection{Total evaluation results from artificially introduced conflicting identities}

\begin{table}[h]
\begin{center}
\begin{tabular}{@{}cc|cccccc@{}}
\toprule
\multirow{2}{*}{\textbf{\(r\)}} & \multirow{2}{*}{\textbf{Trained Method}} & \multicolumn{6}{c}{\textbf{Evaluation Set}} \\
 &  & \textbf{LFW} & \textbf{CALFW} & \textbf{CPLFW} & \textbf{CFP-FP} & \textbf{CFP-FF} & \textbf{AgeDB30} \\ \midrule
     & FaceFusion     & 97.15          & 89.6  & 79.58 & 85.73  & \textbf{97.7}   & 87.38   \\
0\%  & DAIL           & 97.23          & 88.47 & 78.1  & 86.21  & 97.43  & 87.12   \\
     & Naive Concat   & 97.68          & 89.63 & 79.02 & 85.76  & 97.53  & 87.2    \\ \midrule
     & FaceFusion     & 97.55          & \textbf{90.27} & 79.15 & 86.13  & 97.63  & 87.2    \\
20\% & DAIL           & 97.27          & 88.58 & 78.28 & 86.33  & 97.23  & 86.87   \\
     & Naive Concat   & 97.02          & 89.15 & 78.53 & 85.96  & 97.07  & 86.7    \\ \midrule
     & FaceFusion     & 97.6           & 89.78 & 79.77 & \textbf{86.81}  & 97.3   & 87.2    \\
40\% & DAIL           & 96.95          & 89.5  & 78.42 & 86.16  & 97.6   & 86.85   \\
     & Naive Concat   & 96.55          & 89.23 & 78.48 & 86.13  & 97.19  & 86.08   \\ \midrule
     & FaceFusion     & \textbf{97.85}          & 89.82 & \textbf{79.8}  & 86.63  & 97.49  & \textbf{87.75}   \\
60\% & DAIL           & 97.37          & 88.93 & 78.3  & 85.89  & 97.54  & 87.57   \\
     & Naive Concat   & 97.35          & 88.63 & 77.1  & 84.71  & 97.29  & 86.22   \\ \bottomrule     
\end{tabular}
\end{center}
\caption{CASIA Results}\label{table:casia_values}
\end{table}
As shown in Table \ref{table:casia_values}, FaceFusion outperforms other methods regardless of the noise ratio. Table \ref{table:ms1m_values} shows the performance of FaceFusion against all evaluation datasets when trained with artificially divided MS1M dataset, with varying degree of label noises. We are unable to deduce meaningful relationships between the performance of FaceFusion and the noise ratio \(r\), or the presence of duplicated images. This attributes to the robustness of FaceFusion against different compositions of datasets.

\begin{table}[h]
\begin{center}
\begin{tabular}{@{}cc|cccccc@{}}
\toprule
\multirow{2}{*}{\textbf{Trained Method}} & \multirow{2}{*}{\textbf{\(r\)}} & \multicolumn{6}{c}{\textbf{Evaluation Set}} \\
 &  & \textbf{LFW} & \textbf{CALFW} & \textbf{CPLFW} & \textbf{CFP-FP} & \textbf{CFP-FF} & \textbf{AgeDB30} \\ \midrule
 \multirow{6}{*}{w/ image conflicts}  & 0\%  & 99.67 & 95.7  & 89.13 & 93.83 & 99.76 & 97.28   \\
                    & 10\% & \textbf{99.75} & \textbf{95.85} & \textbf{89.97} & 93.7  & 99.73 & 97.37   \\
                    & 20\% & 99.67 & 95.62 & 89.67 & 93.67 & 99.66 & 97.15   \\
                    & 30\% & 99.7  & 95.72 & 89.4  & \textbf{94.03} & 99.74 & 97.2    \\
                    & 40\% & 99.68 & 95.65 & 89.43 & 93.6  & 99.73 & 97.17   \\
                    & 50\% & 99.67 & 95.75 & 89.62 & 93.93 & 99.74 & 97.27   \\ \midrule
\multirow{6}{*}{w/o image conflicts} & 0\%  & 99.67 & 95.7  & 89.13 & 93.83 & 99.76 & 97.28   \\
                    & 10\% & 99.63 & 95.8  & 89.68 & 93.67 & \textbf{99.77} & 97.27   \\
                    & 20\% & 99.68 & 95.73 & 89.27 & 93.99 & \textbf{99.77} & 97.4    \\
                    & 30\% & 99.7  & 95.78 & 89.32 & 93.89 & 99.69 & 97.37   \\
                    & 40\% & 99.68 & 95.7  & 89.82 & 93.96 & 99.66 & 97.25   \\
                    & 50\% & 99.72 & 95.65 & 89.38 & 93.81 & 99.66 & \textbf{97.38}   \\ \bottomrule
\end{tabular}
\end{center}
\caption{MS1M Results}\label{table:ms1m_values}
\end{table}

\subsection{Discussion on in-dataset label noise}
\begin{figure*}[h]
    \begin{center}
    \subfloat[CASIA-WebFace]{\includegraphics[width=.40\linewidth]{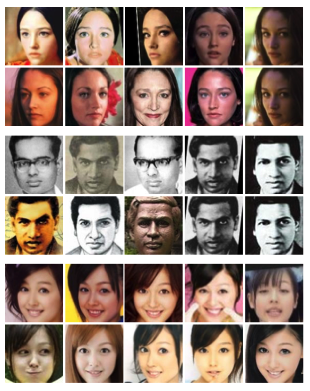}} \label{fig:casia}\quad
    \subfloat[MS1M]{\includegraphics[width=.40\linewidth]{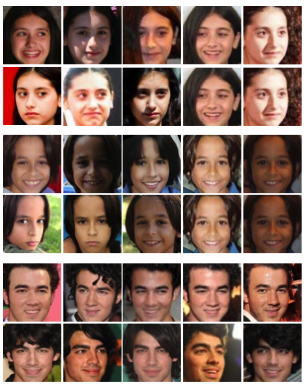} \label{fig:ms1m}}
    \caption{Samples of conflicting identities present in the vanilla datasets. FaceFusion correctly manages these identities by regarding them as if they are conflicting identities from different datasets.}
    \label{fig:samples}
    \end{center}
\end{figure*}

Figure \ref{fig:samples} shows some of the inherent label noise that FaceFusion detects from CASIA, and MS1M datasets. Some of the noisy identities do not consist of the same sets of images. While apparently not the main intention of FaceFusion, the detected label noises shows the capability of FaceFusion in handling identity conflicts. 

\end{document}